\documentclass[runningheads]{llncs}
\usepackage{graphicx}
\usepackage{amsmath}
\usepackage{multirow}
\usepackage{booktabs}
\usepackage{bbding}
\usepackage[colorlinks, linkcolor=blue, anchorcolor=blue, citecolor=blue]{hyperref}
\newcommand{\etal}{\textit{et al.}}
\newcommand{\eg}{\textit{e.g.},}
\newcommand{\ie}{\textit{i.e.},}
\newcommand{\MTHD}{MTHD}
\DeclareMathOperator{\mean}{mean}
\DeclareMathOperator{\var}{var}
\DeclareMathOperator{\mse}{\mathit{MSE}}

\begin{document}
\title{Hetero-Modal Learning and Expansive Consistency Constraints for Semi-Supervised Detection from Multi-Sequence Data}
\titlerunning{Hetero-Modal Learning and Expansive Consistency Constraints}
%
\author{Bolin Lai\inst{1} \and Yuhsuan Wu\inst{1} \and Xiao-Yun Zhou\inst{2} \and Peng Wang\inst{3} \and Le Lu\inst{2} \and Lingyun Huang\inst{1} \and Mei Han\inst{2} \and Jing Xiao\inst{1} \and Heping Hu\inst{3} \and Adam P. Harrison\inst{2}}
\authorrunning{B. Lai et al.}
%
\institute{Ping An Technology, China \\ 
\email{lai.b.bryan@gmail.com} \and
PAII Inc. USA \and
Eastern Hepatobiliary Surgery Hospital, China}

%
\maketitle              
\begin{abstract}
Lesion detection serves a critical role in early diagnosis and has been well explored in recent years due to methodological advances and increased data availability. However, the high costs of annotations hinder the collection of large and completely labeled datasets, motivating semi-supervised detection approaches. In this paper, we introduce mean teacher hetero-modal detection (\MTHD{}), which addresses two important gaps in current semi-supervised detection. \emph{First}, it is not obvious how to enforce unlabeled  consistency constraints across the very different outputs of various detectors, which has resulted in various compromises being used in the state of the art. Using an anchor-free  framework, MTHD formulates a mean teacher approach without such compromises, enforcing consistency on the soft-output of object centers and size. \emph{Second}, multi-sequence data is often critical, \eg{} for abdominal lesion detection, but unlabeled data is often missing sequences. To deal with this, MTHD incorporates hetero-modal learning in its framework. Unlike prior art, MTHD is able to incorporate an expansive set of consistency constraints that include geometric transforms and random sequence combinations.  We train and evaluate MTHD on liver lesion detection using the largest MR lesion dataset to date ($1099$ patients with $>5000$ volumes). MTHD surpasses the best fully-supervised and semi-supervised competitors by $10.1\%$ and $3.5\%$, respectively, in average sensitivity.

\end{abstract}
\section{Introduction}

Lesion detection is a fundamental task in medical imaging, as an end goal~\cite{Castellino_2005} or as a critical step for computer-aided diagnosis (CAD)~\cite{Suzuki_2012}. This puts great impetus on developing powerful lesion detectors and there are many successful deep-learning efforts~\cite{yan20183d,yan2019mulan,huo2020harvesting,jiang2020elixirnet,li2019mvp}. Because of the data-driven nature of deep learning, they rely on a large number of manually annotated images. But annotating bounding boxes (bboxes) is  labor-intensive and time-consuming, requiring roughly $15$ minutes per study~\cite{huo2020harvesting}. A highly promising alternative is to use localization annotations found within hospital picture archiving and communication systems (PACSs)~\cite{yan2018deeplesion}, but such annotations are not typically recorded and are incomplete even when present~\cite{cai2020lesion}. Thus, much like other medical imaging applications~\cite{Tajbakhsh_2020}, lesion detection requires effective techniques for imperfect labels, particularly ones that can best handle unlabeled and heterogeneous PACS data, which is often the only realistic source of large-scale data. 

Most semi-supervised learning (SSL) methods are designed for classification~\cite{berthelot2019mixmatch,laine2017temporal,tarvainen2017mean} or segmentation~\cite{raju2020co,zhou2019collaborative}, and SSL detection is understudied. Unique to detection, its outputs are not easily made consistent, where even the number of predicted bboxes can differ. This has prevented a general framework for enforcing detector consistency. STAC~\cite{sohn2020simple} and unbiased teacher~\cite{liu2021unbiased} offer two good examples, both of which generate pseudo bbox labels to get around this issue. However, pseudo-labels are generated by hard thresholding  detection confidences, which can amplify noise near the margin. Also detection confidence does not necessarily filter good bbox size predictions, which is why unbiased teacher avoids using bbox size consistency~\cite{liu2021unbiased}. Finally, because bboxes are restricted to have horizontal and vertical edges, geometric transforms change their widths and heights, meaning it is actually incorrect to enforce SSL consistency after such transformations. CSD~\cite{jeong2019consistency} avoids it by only enforcing consistency on horizontally flipped counter-parts, but this is a highly limited augmentation scheme. There are successful domain specific techniques, \eg{} for lung nodules~\cite{wang2020focalmix} or fracture regression~\cite{wang2020knowledge}, but these are not generally applicable for lesion detection. 

Another challenge is dealing with multi-sequence data, which is particularly important in abdominal studies~\cite{Burrowes_2017}. Early fusion (EF) is the most straightforward approach, which concatenates sequences together as multi-channel input. However, PACS data is frequently heterogeneous, \ie{} missing sequences/contrast phases~\cite{raju2020co}. Given the expense of annotations, studies with all sequences available are typically chosen for annotation and missing-sequences studies remain unlabeled. EF would abandon such studies, which wastes valuable medical data.

We introduce mean teacher hetero-modal detection (\MTHD{})---an SSL and multi-sequence detector that addresses the above gaps.  We build MTHD off of the anchor-free CenterNet framework~\cite{zhou2019objects}, a leading solution for lesion detection~\cite{cai2020lesion,yan2020learning}. \MTHD{} incorporates hetero-modal learning~\cite{Havaei_2016}, which naturally can handle challenging unlabeled PACS data with missing sequences. Moreover, exploiting CenterNet's heatmap based predictions, \MTHD{} uses a semi-supervised formulation that enforces \emph{valid} consistency across ``centered-ness'' and bbox size, using an \emph{expansive and rich} set of SSL consistency transforms that include intensity, geometric, and hetero-modal sequence combinations. We integrate these constraints within a mean teacher framework~\cite{tarvainen2017mean}. Experiments employ a multi-sequence dataset of $1099$ liver magnetic resonance (MR) studies, $70\%$ of which are unlabeled and heterogeneous. We evaluate on $430$ labeled studies via cross-validation. \emph{This is the largest MR lesion detection study and the largest multi-sequence detection study of any type to date}. When tested on labeled data proportions ranging from $30\%$ to $5\%$ \MTHD{} improves the average sensitivity by $4.2\%$ to $10.8\%$ compared to the best fully-supervised alternative. \MTHD{} also outperforms leading SSL approaches~\cite{laine2017temporal,sohn2020simple,liu2021unbiased} by significant margins, \eg{} $3.5\%$ over the best competitor~\cite{liu2021unbiased}.  

\section{Method}

\begin{figure}[t]
\centering
\includegraphics[width=1.0\columnwidth]{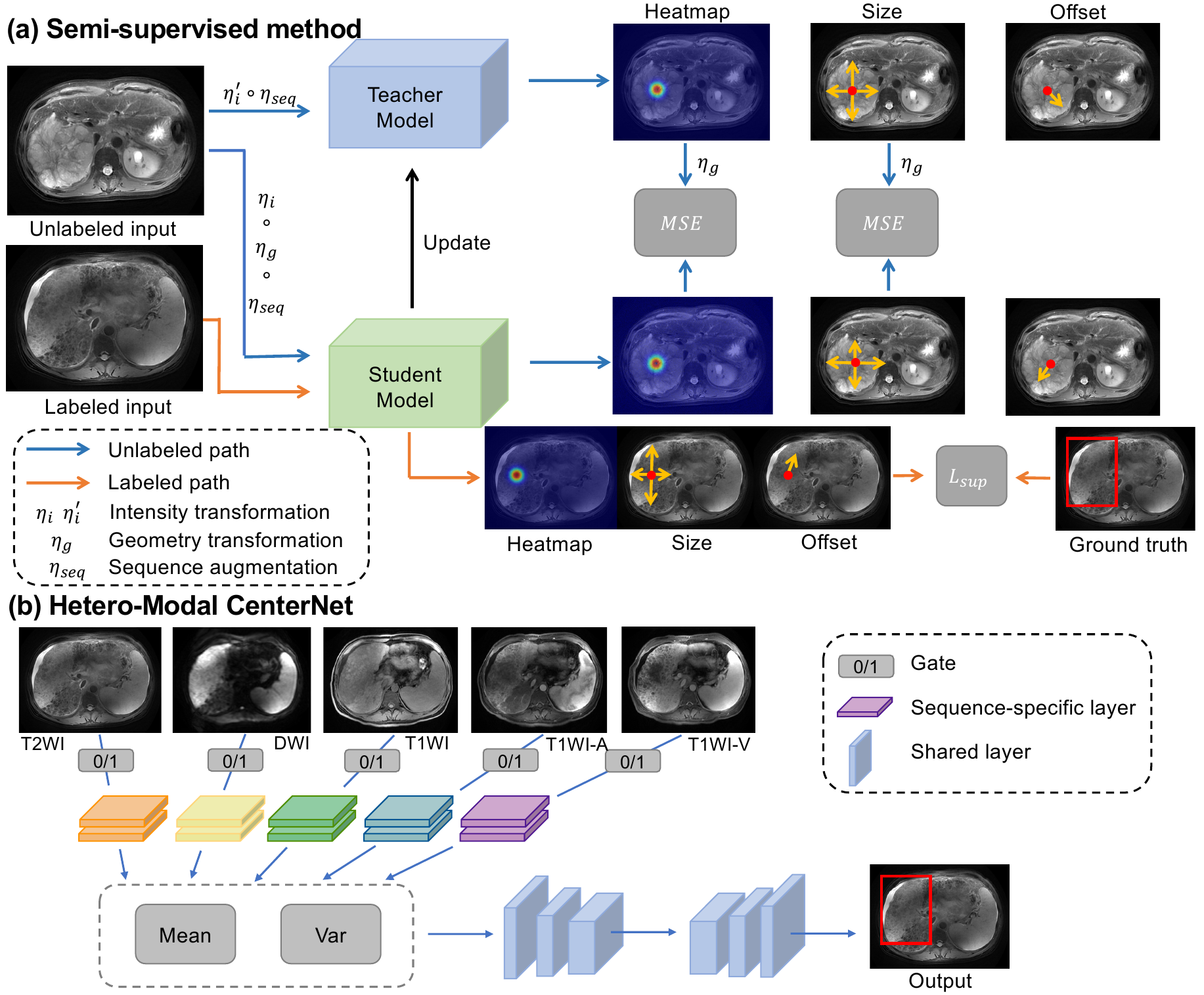}
\caption{The proposed (a) semi-supervised framework and (b) Hetero-Modal CenterNet.}
\label{fig: method}
\end{figure}

Fig.~\ref{fig: method} illustrates the \MTHD{} approach. We assume we are given a multi-sequence dataset consisting of $M$ labeled patients $\mathcal{S}=\{\mathcal{X}_i, \mathcal{Y}_i\}_{i=1}^M$ and $N$ unlabeled patients $\mathcal{U}=\{\mathcal{X}_i\}_{i=1}^N$, where $\mathcal{X}_i$ and $\mathcal{Y}_i$ denote sets of sequences and bbox annotations, respectively. We do not assume that every study has all sequences. This is particularly likely for unlabeled data, which we can divide up into complete and incomplete studies: $\mathcal{U}=\mathcal{U}_{\mathrm{cplt.}} \bigcup \mathcal{U}_{\mathrm{incplt.}}$.

\subsection{Hetero-Modal CenterNet}

Like many SSL approaches, MTHD is built off of a fully-supervised foundation. For reasons elaborated in Sec.~\ref{sec:ssl}, we opt for an anchor-free fully-convolutional network (FCN) detector. While there are several viable options, we choose CenterNet~\cite{zhou2019objects}, which is a leading lesion detector~\cite{cai2020lesion,yan2020learning}. Very briefly, instead of anchors, CenterNet encodes ground-truth locations by splatting a Gaussian kernel at the center of each object onto a heatmap, $Y\in\mathcal{R}^{H/R\times W/R}$ where $R$ is the FCN downsampling factor. A chosen FCN is used to generate a predicted heatmap, $\hat{Y}$, which is encouraged to match $Y$ using a focal-style loss. In addition to this ``centeredness'' heatmap, the FCN also outputs bbox size, $\hat{D}\in\mathcal{R}^{H/R\times W/R \times 2}$, and offset, $\hat{O}\in\mathcal{R}^{H/R\times W/R \times 2}$, activation maps, which are trained using an L1 regression loss, but only on ground-truth center locations. See Zhou \etal{}~\cite{zhou2019objects} for more details. 

To fully exploit all unlabeled data, especially those found within PACSs, a detector must be flexible enough to handle studies with missing sequences. To do this, we adapt the CenterNet architecture for hetero-modal learning~\cite{Havaei_2016}. As Fig.\ref{fig: method}(b) illustrates, if we have $k$ sequences, we construct a set of sequence-specific layers early in the FCN backbone, producing a set of sequence-specific intermediate activation maps  $\mathcal{A}=\{a_{ijk}^1, \ldots a_{ijk}^k\}$. Here we assume 2D inputs to create 3D activations. Since some sequences may be missing, this set of activation maps should be aggregated with any operator that can accept a variable number of inputs. As argued by Havaei \etal{}~\cite{Havaei_2016}, the mean and variance are excellent choices, since their expected values should be constant across different numbers of inputs. If the output activation is denoted $b_{ijk}$, this can be expressed as
\begin{align}
    b_{ijk} = \mean(\mathcal{A}) \oplus \var(\mathcal{A}) \textrm{,} \label{eqn:hetero}
\end{align}
where $\oplus$ as channel-wise concatenation. After \eqref{eqn:hetero}, a shared set of layers processes $b_{ijk}$ just like any other standard FCN. During training, random combinations of sequences are chosen, while in inference all available sequences are used to produce the most confident prediction. In addition to being able to handle heterogeneous learning, hetero-modal learning can act as a powerful form of data augmentation. As we show later, this can even boost performance in fully-supervised setups when all sequences are available. Additionally, hetero-modal learning can provide another form of SSL consistency transformation.

\subsection{Semi-Supervised Detection}
\label{sec:ssl}

As shown in Fig.\ref{fig: method}(a), the proposed semi-supervised detector is based off of the popular mean teacher framework~\cite{tarvainen2017mean}. The labeled images, $\mathcal{S}$, are first used to train a hetero-modal CenterNet, denoted $f_{\theta}(\mathcal{X}_{i})$. The resulting network weights are used to initialize teacher and student parameters, $\theta_{t}$ and $\theta_{s}$, respectively. The student model is then trained by transforming unlabeled data via strong augmentations, denoted $\eta$, and penalizing any inconsistencies with the teacher model:
\begin{align}
    \mathcal{L} = \sum_{\{\mathcal{X}_{i},\,\mathcal{Y}_{i}\}\in \mathcal{S}}\ell_{sup}(f_{\theta_{s}}(\mathcal{X}_{i}), \mathcal{Y}_{i}) + \lambda \sum_{\mathcal{X}_{i}\in \mathcal{U}}\ell_{cons}(f_{\theta_{s}}(\mathcal{X}_{i},\eta),f_{\theta_{t}}(\mathcal{X}_{i},\eta')) \mathrm{,}
\end{align}
where $\ell_{sup}(.)$ is the standard CenterNet loss~\cite{zhou2019objects}, $\ell_{cons}(.)$ is some consistency loss, and $f_{\theta}(.)$ has also been modified to accept the augmentations. Only the student model is updated by backpropagation. On the other hand, the ``mean teacher'' parameters are updated using  a moving average of the current student parameters:
\begin{align}
\theta_{t}^k &= \alpha\theta_{t}^{k-1} + (1-\alpha)\theta_{s}^k \mathrm{,}
\end{align}
where  $\alpha$ denotes the moving average rate.

The challenge with SSL detection is how to construct the strong data augmentations and a consistency loss. To avoid difficulties with enforcing consistency across detection outputs, prior student/teacher approaches have relied on using teacher pseudo labels~\cite{sohn2020simple,liu2021unbiased}, but this only allows for hard labels and requires filtering out spurious teacher outputs using a confidence threshold, which does not necessarily filter out poor bbox size predictions~\cite{liu2021unbiased}. Moreover, geometric augmentations are not easily applied to bboxes, since a rotation changes their shape.  Liu \etal{}~\cite{liu2021unbiased} avoid geometric transforms in their mean teacher approach, but given the importance of well-chosen data augmentations~\cite{tarvainen2017mean}, this is critical limitation. Here is where the adoption of an anchor-free FCN-style framework, like CenterNet~\cite{zhou2019objects}, provides important benefits. For one, unlike anchor-based frameworks, consistency can be straightforwardly enforced on FCN outputs, avoiding the complexities of generating pseudo-labels and allowing for soft constraints, similar to what is done in SSL classification and segmentation. Moreover, CenterNet's Gaussian kernel-based heatmap, $\hat{Y}$, is equivariant with rotations, meaning such geometric transforms are not problematic. This simple insight can allow for important performance gains. 

Additionally, because MTHD is hetero-modal, it can apply missing sequence augmentations, $\eta_{seq}$, in addition to intensity and geometric ones, $\eta_{i}$ and $\eta_{g}$. Like Liu \etal{}~\cite{liu2021unbiased}, we only apply ``weak'' transformations  to the teacher, which in our case foregoes geometric transformations.  Instead we transform the \emph{output} of the teacher model with the same geometric transform used on the \emph{input} of the student model. This scheme avoids the danger of a geometric transform occluding a bbox in the teacher input that is visible to the student, which would produce misleading supervisory signals. Formulaically, this is expressed as
\begin{align}
    \hat{Y}_{s}, \hat{D}_{s}, \hat{O}_{s} &= f_{\theta_{s}}(\eta_{i} \circ \eta_{g} \circ \eta_{seq} \circ \mathcal{X}_{i})  \mathrm{,} \\
    \hat{Y}_{t}, \hat{D}_{t}, \hat{O}_{t} &=  f_{\theta_{t}}(\eta_{i}' \circ \eta_{seq} \circ \mathcal{X}_{i}) \mathrm{,} \label{eqn:cons_2}\\
    \ell_{cons} &= \mse(\hat{Y}_{s},\, \eta_{g} \circ \hat{Y}_{t}) + \lambda \mse(\hat{D}_{s},\, \eta_{g} \circ \hat{D}_{t}) \mathrm{,}
\end{align}
where consistency is enforced using the mean-squared error (MSE). Here, the student and teacher use the same set of random sequences, $\eta_{seq}$, because we found training was unstable otherwise. Note that no consistency is used on offset, $\hat{O}$, because it is a local refinement, which is quite different before and after geometric transforms, so it is not correct to enforce consistency across those outputs. 

\section{Results}

\subsubsection{Data and Setup}

We tested MTHD on liver lesion detection from multi-sequence MR studies. We collected $1099$ studies, each ideally containing T1-weighted imaging (T1WI), T2-weighted imaging (T2WI), venous-phase T1WI (T1WI-V), arterial-phase T1WI (T1WI-A), and diffusion-weighted imaging (DWI) MR sequences, from \emph{Anonymized}. The use of multi-sequence data is critical for clinical detection and characterization~\cite{aube2017easl}. Any patients with both radiological and pathological reports between 2006 and 2019 were included in the dataset, $430$ of which, with all five sequences available, were annotated with 2D bboxes on each slice under the supervision of a hepatic physician with $>10$ years experience. Of the remaining $669$ unlabeled studies, $495$ have all five sequences ($\mathcal{U}_{\mathrm{cplt.}}$) and $174$  were missing at least one sequence ($\mathcal{U}_{\mathrm{incplt.}}$). Labeled data was split using five-fold cross validation with $70\%$, $10\%$ and $20\%$ as training, validation, and test sets, respectively. The unlabeled data was included in the training set of each fold.  Gamma transformation was used as the intensity transformation ($\eta_{i}$), and random rotations, resizing, shifting were adopted as geometric transformations ($\eta_{g}$). The official implementation and hyper-parameters of 2D CenterNet~\cite{zhou2019objects} were used in all experiments. Since up to five sequences are inputted to the model at the same time, 3D models consume too much GPU memory for practical use. A listing of hyperparameters can be found in the supplementary.

For SSL, $\alpha$ and $\lambda$ were set as $0.999$ and $0.02$, respectively. We measured performance using the average sensitivity of different numbers of false-positives (FPs) \emph{per patient}, which as Cai \etal{} argue~\cite{cai2020lesion}, is a more meaningful and challenging metric than the \emph{per-slice} numbers often used in DeepLesion works. A prediction bounding box is seen as true positive when IoU$>$0.5. 

\begin{figure}[t]
\centering
\includegraphics[width=0.5\columnwidth]{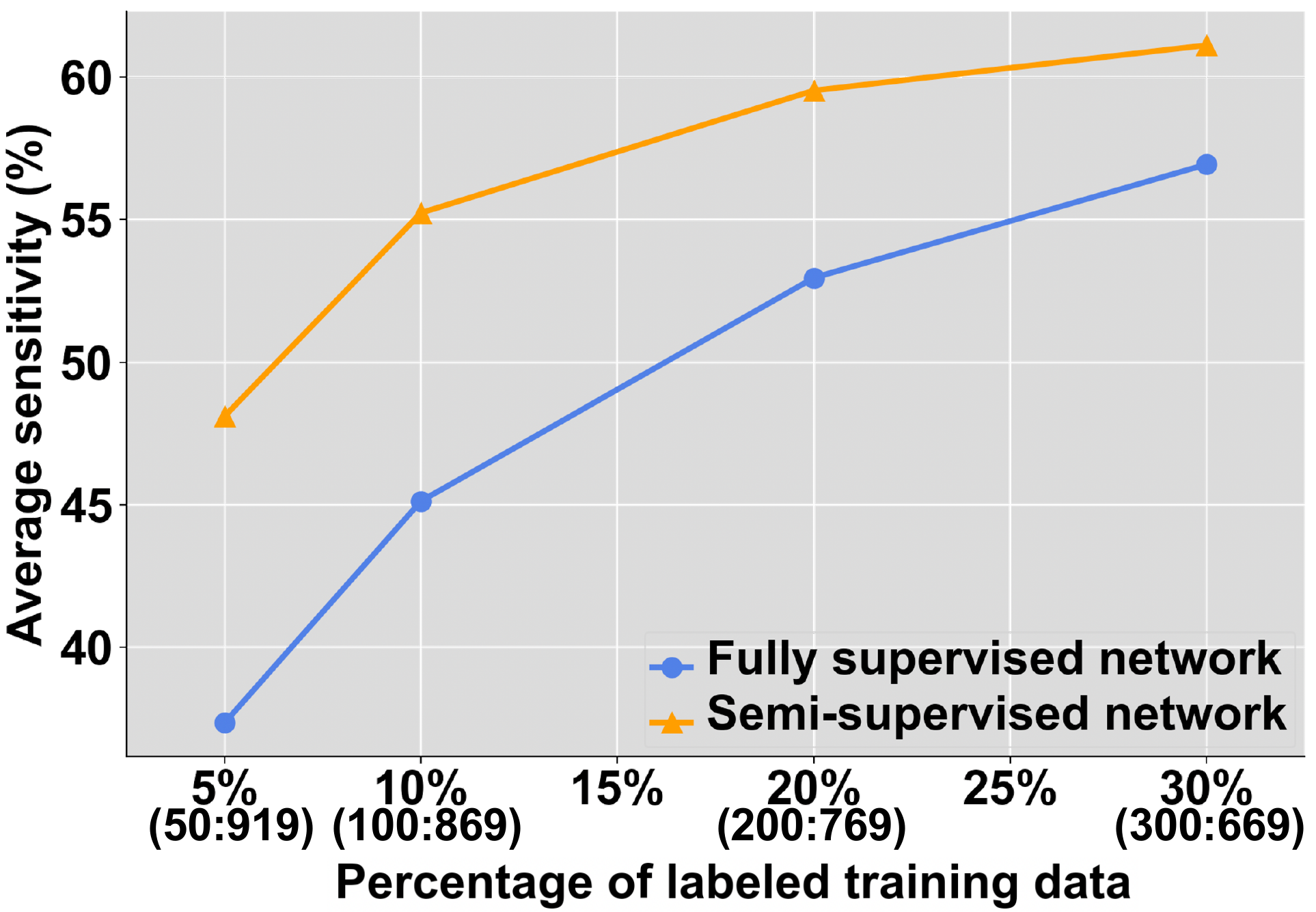}
\caption{Average sensitivity of MTHD compared to the fully-supervised Hetero-Modal CenterNet across different ratios of labeled to unlabeled data.}
\label{fig:percentage}
\end{figure}

\subsubsection{Evaluation of Semi-Supervised Learning}

We first evaluate the improvement of MTHD over Hetero-Modal CenterNet using different percentages of labeled training data. To do this, we randomly treat different proportions of labeled data as unlabeled, keeping the total number of training studies as $969$. The results are shown in Fig.\ref{fig:percentage}, where consistent gains can be observed across all ratios of labeled to unlabeled data, with greater improvements at smaller ratios, \ie{} the improvements at $5\%$, $10\%$, $20\%$ and $30\%$ ratios are $10.8\%$, $10.1\%$, $6.6\%$ and $4.2\%$, respectively. This validates MTHD's ability to effectively use unlabeled data to provide significant boosts in performance.

\subsubsection{Comparison with State of the Art}

\begin{table}[t]
\caption{Comparison against state of the art. In all experiments, the results are average of five-fold cross validation on test set with 10\% training data labeled.}\label{tab:state-of-the-art}
\setlength{\tabcolsep}{2.5mm}
\centering
\begin{tabular}{lcccccc}
\toprule
\multirow{2}{*}{Methods}  & \multicolumn{5}{c}{FPs per patient}  & \multirow{2}{*}{Avg.}\\
\cmidrule(lr){2-6}
& 1 & 2 & 4 & 8 & 16 & \\
\midrule
\multicolumn{7}{c}{\emph{Fully-supervised models}} \\
\midrule
3DCE~\cite{yan20183d}                     & 17.3 & 22.6 & 29.3 & 36.8 & 43.0 & 29.8 \\
CenterNet~\cite{zhou2019objects}          & 23.2 & 32.1 & 42.0 & 52.0 & 59.3 & 41.9 \\
ATSS~\cite{zhang2020bridging}             & 25.4 & 31.0 & 37.6 & 44.6 & 50.2 & 37.7 \\
Hetero-Modal CenterNet                    & 25.8 & 35.6 & 45.8 & 55.6 & 62.9 & 45.1 \\
\midrule
\multicolumn{7}{c}{\emph{Semi-supervised models}} \\
\midrule
Co-training~\cite{raju2020co}             & 17.3 & 22.3 & 28.8 & 36.1 & 43.2 & 29.5 \\
$\rm{\Pi}$-Model~\cite{laine2017temporal} & 29.4 & 38.4 & 47.5 & 56.4 & 64.6 & 47.3 \\
CSD~\cite{jeong2019consistency}           & 29.8 & 37.6 & 45.7 & 54.4 & 62.7 & 46.1 \\
STAC~\cite{sohn2020simple}                & 30.5 & 40.5 & 50.0 & 57.6 & 63.0 & 48.3 \\
Unbiased Teacher~\cite{liu2021unbiased}   & 32.1 & 42.8 & 53.3 & 62.0 & 68.3 & 51.7 \\
\midrule
\MTHD{}                                   & \textbf{37.0} & \textbf{47.6} & \textbf{56.3} & \textbf{64.5} & \textbf{70.8} & \textbf{55.2} \\
\bottomrule
\end{tabular}
\end{table}

We compare \MTHD{} with state of the art in Table~\ref{tab:state-of-the-art}. First, we compare Hetero-Modal CenterNet with three early fusion (EF) detectors. Even though 3DCE~\cite{yan20183d} and ATSS~\cite{zhang2020bridging} are strong baselines for DeepLesion~\cite{yan2018deeplesion} and COCO~\cite{lin2014microsoft}, respectively, CenterNet~\cite{zhou2019objects} outperforms them by a large margin, which validates its reported effectiveness on lesion detection~\cite{cai2020lesion,yan2020learning}.  Hetero-Modal CenterNet can boost performance even further, producing an average sensitivity of $45.1\%$, surpassing its EF counter-part by $3.2\%$. This occurs even though inference is always performed using all five MR sequences. We postulate that by randomly selecting sequences during training, hetero-modal learning can provide a form of data augmentation that benefits even non-heterogeneous inference.

We compare against several SSL methods, most of which follow a student/teacher framework~\cite{laine2017temporal,jeong2019consistency,sohn2020simple,liu2021unbiased}. Because we have multi-sequence data, we also compare against a co-training~\cite{Blum_1998} implementation that follows Raju \etal{}'s approach~\cite{raju2020co}. \emph{To conduct a fair comparison, we use Hetero-Modal CenterNet as the detector framework for all options}. Among prior work, unbiased teacher~\cite{liu2021unbiased} achieves the best performance. This is not surprising because, like MTHD, it uses the mean teacher framework, which is a highly effective SSL approach. However, unbiased teacher relies on pseudo labels that are generated by hard thresholding detection confidences. This means that any inaccuracies in the teacher output near the threshold will be amplified by polarizing the labels to hard positives or negatives. Moreover, unbiased teacher does not apply a consistency loss to the bbox size regression because thresholding the detection confidence does not filter for good width and height predictions~\cite{liu2021unbiased}. In contrast, \MTHD{} uses a consistency loss to penalize differences between student and teacher outputs on both center locations and bounding box sizes. This can be seen as a kind of soft pseudo label, which does not amplify noise in the teacher outputs. The significant improvement of $3.5\%$ validates its superiority. Visual examples, seen in the supplementary, provides further qualitative validation. 

\subsubsection{Ablation Study}

\begin{table}[t]
\caption{Ablation study of MTHD. Results correspond to the average of validation results across the five folds using the 10\% labeled training data setting.}\label{tab:ablation}
\setlength{\tabcolsep}{0.8mm}
\centering
\begin{tabular}{lcccccccc}
\toprule
\multirow{2}{*}{Methods} & Intensity & Geometric & \multicolumn{5}{c}{FPs per study} & \multirow{2}{*}{Avg.} \\
\cmidrule(lr){4-8}
 & trans.    & trans.   & 1 & 2 & 4 & 8 & 16 & \\
\midrule
2.5D Hetero-Modal CenterNet           & N/A     & N/A        & 26.3 & 33.0 & 41.0 & 47.8 & 53.9 & 40.4 \\
2D Hetero-Modal CenterNet             & N/A     & N/A        & 24.9 & 34.8 & 45.0 & 53.4 & 60.5 & 43.7 \\
2D+center consistency                 & $\surd$ & \texttimes & 32.9 & 39.2 & 46.0 & 53.1 & 58.9 & 46.0 \\
2D+center consistency                 & $\surd$ & $\surd$    & 38.8 & 46.1 & 54.1 & 62.1 & 67.7 & 53.8 \\
\MTHD                                 & $\surd$ & $\surd$    & $\mathbf{39.0}$ & $\mathbf{47.1}$ & $\mathbf{56.0}$ & $\mathbf{63.0}$ & $\mathbf{69.1}$ & $\mathbf{54.9}$ \\
\MTHD{} w/o $\mathcal{U}_{\mathrm{incplt.}}$  & $\surd$ & $\surd$ & 34.9 & 41.0 & 47.4 & 53.5 & 58.6 & 47.1 \\
\bottomrule
\end{tabular}
\end{table}

Table~\ref{tab:ablation} presents an ablation study. In terms of fully-supervised backbone, we also tried a 2.5D variant, seen in other works~\cite{cai2020lesion}, but its performance was inferior to the 2D version. This may be due to the large slice thickness (8mm) of raw MR data. Once unlabeled data is included, it can be seen that enforcing consistency on the centeredness heatmap, $\hat{Y}$, improves the average sensitivity by $2.3\%$ when only using intensity transformations. Geometric transformations further boost the score from $46.0\%$ to $53.8\%$, highlighting how critical strong data augmentations are to mean teacher frameworks. Adding consistency on bounding-box sizes then completes MTHD, which adds another $1.1\%$ improvement. Finally, to demonstrate the importance of using all unlabeled data, we train MTHD without studies having missing sequences ($\mathcal{U}_{\mathrm{incplt.}}$), which results in a considerable decrease in performance. This underscores that using all available data, even if imperfect, can be critical for performance. As such, this further validates our use of hetero-modal learning as a foundation for SSL detection on multi-sequence data. 

\section{Conclusion}

Lesion detection greatly benefits from having large-scale data. Nonetheless, given the cost of manual annotations, SSL is a crucial strategy to best exploit clinical data archives. Towards this end, we articulate a hetero-modal SSL detection technique, called MTHD, that (1) uses hetero-modal learning to handle missing sequences and to exploit all available unlabeled data; and (2) employs expansive consistency constraints for much more effective SSL. Using the largest MR liver detection dataset to date, MTHD improves the average sensitivity by up to $10\%$ compared to fully-supervised counter-parts and can outperform leading SSL alternatives by $3.5\%$ to $9.1\%$. As such, MTHD represents an important step forward to SSL lesion detection.

%
\bibliographystyle{splncs04}
\bibliography{mybibliography}

\newpage
~\\
\centerline{\LARGE \textbf{Supplementary Material}}

\setcounter{section}{0}  

\section{Implementation Details}

\subsubsection{Preproessing} In each study, images of all sequences were resampled to $7.58mm\times7.58mm\times2.67mm$ and registered to T2WI with DEEDS algorithm~\cite{gletsos2003computer}. We also clipped them using 0.1\% and the 99.9\% intensity percentile values.

\subsubsection{Training} The fully-supervised Hetero-Modal CenterNet was trained using Adam optimizer~\cite{Kingma2015AdamAM} with a learning rate of $1.25\times10^{-4}$ and batch size of 30 for 30 epochs. The weight decay was $1\times10^{-5}$. The backbone used in all experiments for CenterNet~\cite{zhou2019objects} was ResNet101~\cite{he2016deep} The number of slices with tumors and slices without tumors in each batch was 2:1. All other setups were the same as official protocols. The combination inputted into the network was chosen by first randomly choosing the number of sequences and then randomly choosing one combination under this number. In semi-supervised learning, the learning rate was reduced to $5\times10^{-5}$. Other setups for labeled data were kept unchanged. As for unlabeled data, different gamma transformation ([0.5, 2]) was added as intensity transformation to teacher and student inputs. Extra geometric transformations were implemented to student input, including random rotation ($[-10^\circ, +10^\circ]$), random resizing ([0.8, 1.25]), and random shifting ($[-\frac{H}{4}, +\frac{H}{4}]$ and $[-\frac{W}{4}, +\frac{W}{4}]$). Like the fully-supervised setups, we inputted random hetero-modal sequence combinations, but used the same choice for both teacher and student. The batch size was set to 12 and the ratio of labeled images and unlabeled images was 1:1.

\section{Visualization of Detection Results}

\begin{figure}[t]
\centering
\includegraphics[width=1.0\columnwidth]{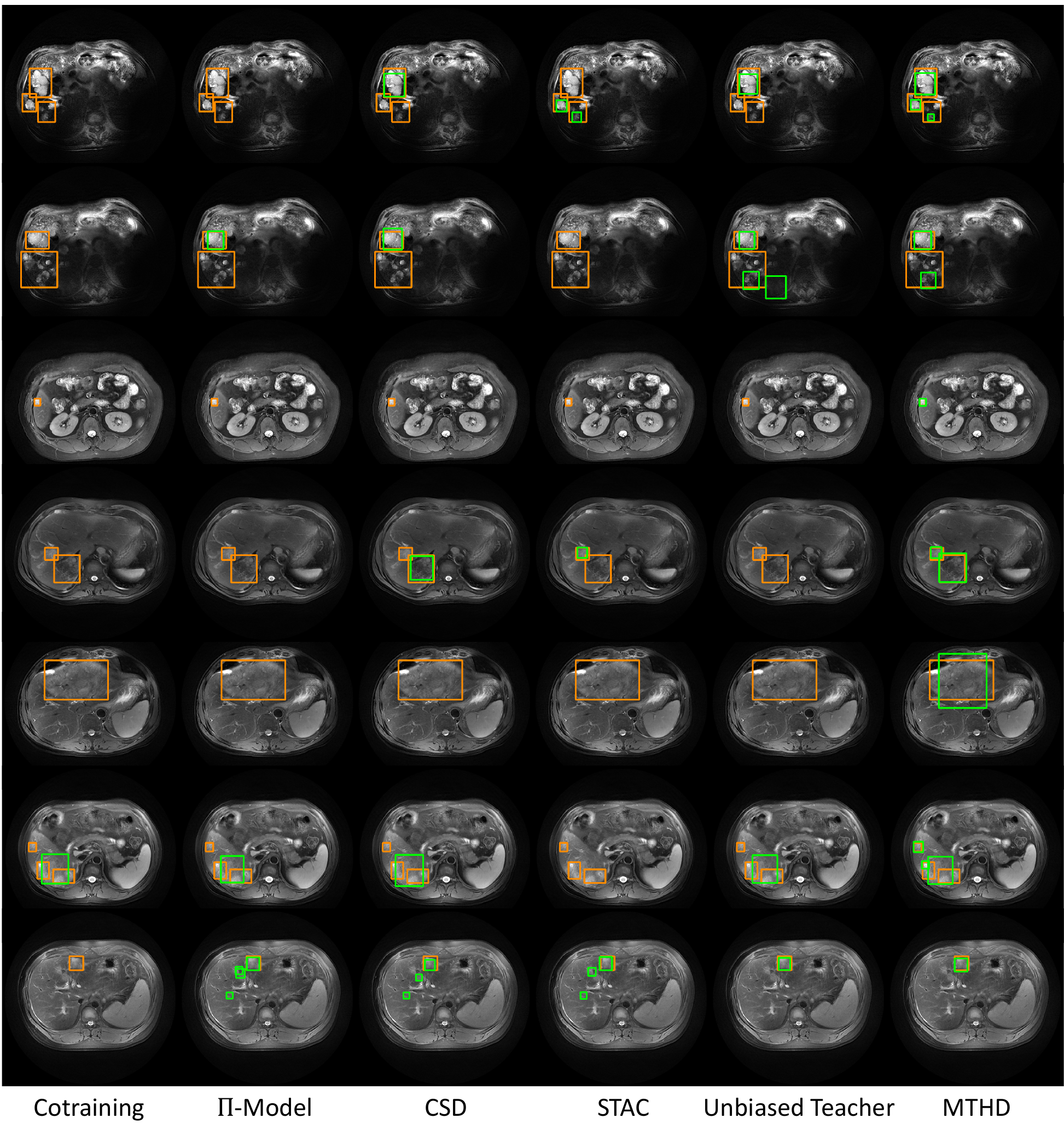}
\caption{Visualization of detection results across our method and other baselines of semi-supervised detection when there are 4 FPs in each patient on average. Green and orange bounding boxes are predictions and ground truth, respectively.}
\label{fig:visualization}
\end{figure}

Fig.\ref{fig:visualization} demonstrates the comparison across different semi-supervised detection methods. To make a fair comparison, the threshold of each method is independently chosen to keep 4 FPs per patient on average. The baseline methods are very likely to miss some tumors or generate more FPs especially when tumors are very small or distribute in a cluster. Additionally, the size of bounding boxes from MTHD is more close to ground truth, which further validates the superiority of the proposed method.

\end{document}